%% file: main.tex
\documentclass[final,3p,times,onecolumn]{elsarticle}
\usepackage{setspace}
\input{0-2-packages_and_commands}
\input{0-3-acronyms}

\makeatletter
\def\ps@pprintTitle{%
}
\makeatother

\usepackage{tikz}
\usepackage{transparent}
\usepackage[hidelinks=true,colorlinks=true,allcolors=black]{hyperref}

\newcommand\copyrighttext{%
  \scriptsize This is an author-prepared version. The published version is available on \textit{ScienceDirect} (DOI: \href{http://doi.org/10.1016/j.jvcir.2024.104385}{10.1016/j.jvcir.2024.104385}).}
\newcommand\copyrightnotice{%
\begin{tikzpicture}[remember picture,overlay]
\node[anchor=south,yshift=105pt,xshift=0pt] at (current page.south) {\fbox{\transparent{0.85}\parbox{\dimexpr0.745\textwidth-\fboxsep-\fboxrule\relax}{\copyrighttext}}};
\end{tikzpicture}%
}

\begin{document}
    \input{0-1-frontmatter}
    \modulolinenumbers[5]
    \copyrightnotice
    \glsresetall
    \doublespacing
    \input{1-introduction}
    \input{2-related}
    \input{3-methodology}
    \input{4-protocol}
    \input{5-results}

    \input{7-conclusion}
    \input{9-acknowledgments}
    \bibliographystyle{model2-names}
    \bibliography{references}
\end{document}

%% file: 0-2-packages_and_commands.tex
\usepackage[numbers]{natbib}
\usepackage{lineno}
\usepackage{times}
\usepackage{epsfig}
\usepackage{graphicx}
\usepackage{amsmath}
\usepackage{amssymb}
\usepackage{multirow}
\usepackage[caption=false,farskip=0pt]{subfig}

\usepackage{pdflscape}
\usepackage{graphicx,caption,rotating}
\usepackage{setspace}
\usepackage{lscape}

\usepackage[hyphens]{url}

\usepackage{balance}
\usepackage{booktabs}
\usepackage[space]{cite}
\usepackage[acronym]{glossaries}
\usepackage{xcolor}
\usepackage{soul}
\usepackage[normalem]{ulem} 

\definecolor{violet}{rgb}{0.53, 0.0, 0.69}

\newcommand\tbf[1]{\textbf{#1}}

\newcommand\minor[1]{#1} 

\usepackage{pifont}
\newcommand{\cmark}{\ding{51}}%
\newcommand{\xmark}{\ding{55}}%
\newacronymstyle{long-short-br}
{%
  \GlsUseAcrEntryDispStyle{long-short}%
}%
{%
  \GlsUseAcrStyleDefs{long-short}%
}
\setacronymstyle{long-short-br}

%% file: 0-3-acronyms.tex
\newacronym{bisst}{Bi-SST}{bidirectional single-stream}
\newacronym{bmt}{BMT}{Bi-Modal Transformer}
\newacronym{cnn}{CNN}{Convolutional Neural Network}
\newacronym{dap}{DAP}{Deep Action Proposal}
\newacronym{dvc}{DVC}{Dense Video Captioning}
\newacronym{glove}{GloVE}{Global Vectors}
\newacronym{gru}{GRU}{Gated Recurrent Unit}
\newacronym{lstm}{LSTM}{Long Short-Term Memory}
\newacronym{mdvc}{MDVC}{Multi-modal Dense Video Captioning}
\newacronym{mse}{MSE}{Mean Squared Error}
\newacronym{nlp}{NLP}{Natural Language Processing}
\newacronym{of}{OF}{Optical Flow}
\newacronym{rnn}{RNN}{Recurrent Neural Network}
\newacronym{sota}{SOTA}{state-of-the-art}
\newacronym{tac}{TAC}{trimmed action classification}

%% file: 0-1-frontmatter.tex
\begin{frontmatter}
\title{Dense Video Captioning Using Unsupervised Semantic Information}

\author[1,2]{Valter Estevam\corref{mycorrespondingauthor}}
\cortext[mycorrespondingauthor]{Corresponding author}
\ead{valter.junior@ifpr.edu.br}
\author[2,3]{Rayson Laroca}
\author[4]{Helio Pedrini}
\author[2]{David Menotti}

\address[1]{Federal Institute of Paran\'a, Irati-PR, 84500-000, Brazil}
\address[2]{Federal University of Paran\'a, Department of Informatics, Curitiba-PR, 81531-970, Brazil}
\address[3]{Pontifical Catholic University of Paran\'a, Postgraduate Program in Informatics, Curitiba-PR, 80215-901, Brazil}
\address[4]{University of Campinas, Institute of Computing, Campinas-SP, 13083-852, Brazil\\[3ex]\texttt{vlejunior@inf.ufpr.br}\qquad\texttt{rayson@ppgia.pucpr.br}\qquad\texttt{helio@ic.unicamp.br}\qquad\texttt{menotti@inf.ufpr.br}\\[-5ex]}


\begin{abstract}
We introduce a method to learn unsupervised semantic visual information based on the premise that complex events can be decomposed into simpler events and that these simple events are shared across several complex events. We first employ a clustering method to group representations producing a visual codebook. Then, we learn a dense representation by encoding the co-occurrence probability matrix for the codebook entries. This representation leverages the performance of the dense video captioning task in a scenario with only visual features. For example, we replace the audio signal in the BMT method and produce temporal proposals with comparable performance.  Furthermore, we concatenate the visual representation with our descriptor in a vanilla transformer method to achieve state-of-the-art performance in the captioning subtask compared to the methods that explore only visual features, as well as a competitive performance with multi-modal methods. Our code is available at \url{https://github.com/valterlej/dvcusi}.
\end{abstract}
\end{frontmatter}

%% file: 1-introduction.tex
\section{Introduction}

\glsresetall

In this work, we aim to perform \gls{dvc}~\citep{krishna:2017} using only visual features.
\gls{dvc} is a complex task that involves \minor{localizing} events and providing a suitable description for them in untrimmed videos. \minor{This task differs from Video Captioning because the events are usually not perfectly delimited, making generating accurate captions more challenging.} \minor{Nowadays, \gls*{dvc}} has been tackled using multi-modal features: visual and audio~\citep{iashin:2020b}, visual, audio, and speech~\citep{iashin:2020,chadha:2021}. 
\minor{Nevertheless, audio features may not always be available or indicative of the content in the video, and the same holds true for speech features.}
\minor{Therefore, it becomes imperative to devise approaches that rely solely on visual information.}
In this sense, we propose a new visual descriptor learned with an unsupervised method that can encode the co-occurrence visual similarity of short video clips (i.e., lasting a few seconds) to be used in the \gls*{dvc} task. Our inspiration is that humans can recognize similar video fragments and infer the later scenes from a movie they have not seen before, relying entirely on their prior knowledge and contextual information.

Recently, several methods have been proposed for learning deep representations in an unsupervised manner~\citep{xie:2016,hsu:2018,caron:2018,huang:2020}. These methods usually combine a deep neural network (e.g., CNN or autoencoders) and a clustering method (e.g., $k$-means or agglomerative clustering). In the general framework, clusters are used to organize latent representations into soft labels which, in turn, are used in a supervised model that updates the encoder weights~\citep{aljalbout:2018}, improving the latent features. However, our goal is slightly different. We are interested in generating a dense representation encoding the visual relationships, where short clips are similar to each other and occur in their temporal context. These relationships are not captured by the aforementioned methods, which are optimized to produce more discriminative features.

\begin{figure}[htb!]
\captionsetup{captionskip=0.75pt}
\centering
\subfloat[][\label{fig:similarityillustration_a}]{
\includegraphics[width=0.75\linewidth]{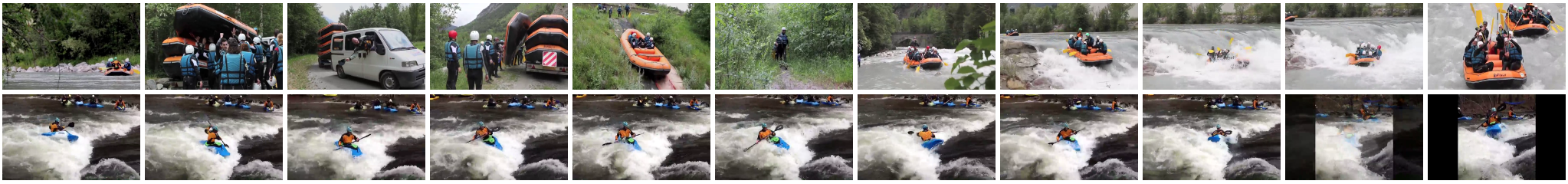}}%

\vspace{2.5mm}

\subfloat[][\label{fig:similarityillustration_b}]{
\includegraphics[width=0.75\linewidth]{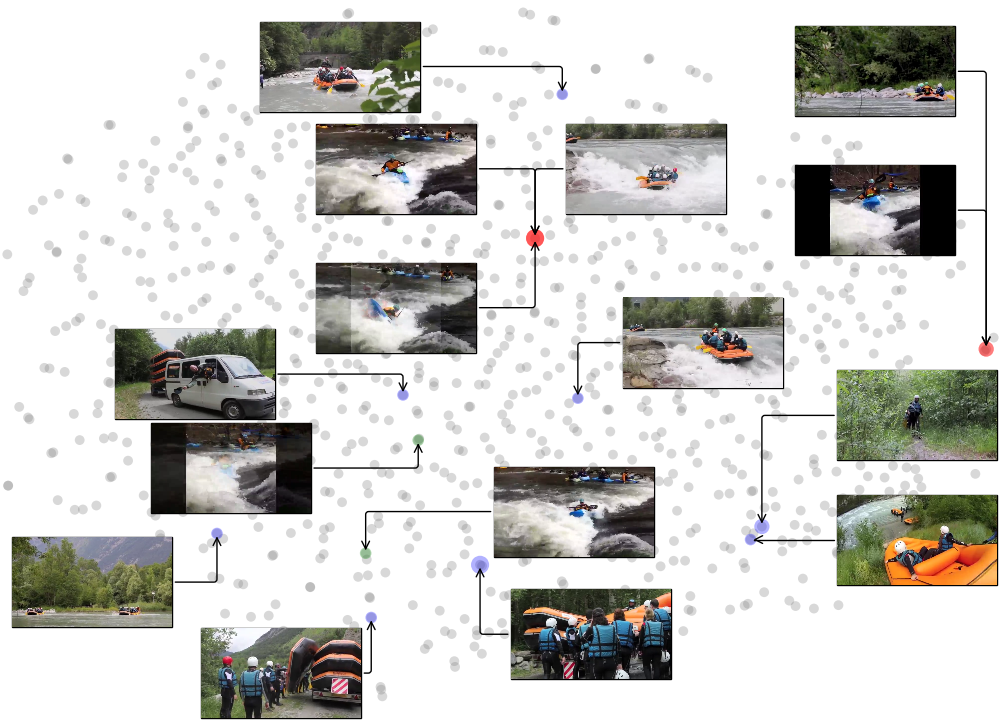}}%

\vspace{-1mm}

\caption{
Examples of visual similarities.
(a) Two video fragments with about 28 seconds from YouTube (v\underline{\space}dBNZf90PLJ0 and v\underline{\space}j3QSVh\underline{\space}AhDc). They share some visual similar short clips. (b) A 2D t-SNE representation for the whole visual vocabulary. Some shared fragments are highlighted in red.}
\label{fig:similaryillustration}
\end{figure}

The \minor{intuition} behind the proposed method is that long and complex events can be decomposed into short and simple events, as illustrated in Figure~\ref{fig:similarityillustration_a} --~which shows two videos of related water sports: rafting and kayaking.
We first identify similar events by splitting the videos into short clips and then extract visual features using the i3D method~\citep{carreira:2017} for each short clip.
A mini-batch $k$-means method groups representations based on their Euclidean distance producing a visual codebook, and a discrete representation is obtained by the sequence of cluster label numbers. 
Afterward, inspired by the GloVe method~\citep{pennington:2014}, we compute a co-occurrence matrix for this codebook and learn a dense representation by training a neural network to predict the pre-computed co-occurrence probability of any two visual words, as detailed in Section~\ref{sec:cooccurence}.

In Figure~\ref{fig:similarityillustration_b}, a 2D t-SNE~\citep{maaten:2008} visualization was used to project the entire visual codebook (drawn as gray dots).
The clips from the first and second videos are represented with blue and green dots, respectively.
Observe that the final content from the first video is much similar to the content of the second one (see the red dots) and that fragments with similar content are close to each other.

Our semantic descriptor can be employed in the \gls{dvc} task, which consists of two subtasks: temporal event proposal and video captioning.
In this work, we employ a popular strategy of handling these tasks independently.
More specifically, we use a multi-headed bi-modal proposal module~\citep{iashin:2020b} for event proposal generation, and a vanilla transformer~\citep{vaswani:2017} for video~captioning.


In summary, the contributions of this work are (i)~an unsupervised descriptor that can be easily employed in dense video captioning\minor{, using either visual features alone or in combination with multiple modalities}; (ii)~the visual similarity proved to be efficient to generate event proposals replacing the audio signal adopted in \minor{the BMT method}~\citep{iashin:2020b}; and (iii)~our captioning results in \minor{the learned proposals scenario (i.e., the most complex setting) showcase the descriptor's ability to accurately capture the visual similarity between both seen and unseen clips}, achieving state-of-the-art performance \minor{when} considering only visual features and competitive performance \minor{when} compared to multi-modal~methods.

%% file: 2-related.tex
\section{Related work}


Dense video captioning was first introduced in~\citep{krishna:2017}.
It involves proposing a temporal event localization in untrimmed videos (i.e., event proposal generation) and providing a suitable description for the event in fluent natural language (i.e., video captioning). \minor{Presently, \gls*{dvc} research has primarily focused on general events~\citep{krishna:2017}, sport events~\citep{yu:2018}, and cooking activities~\citep{zhou:2018}.
For this particular study, our focus lies on general events.
In Section~\ref{subsec:eventproposalgenerationref}, we elaborate on the event proposal generation, while Section~\ref{subsec:videocaptioningref} is dedicated to discussing the video captioning subtask.
In Section~\ref{subsec:unsuperviseddeeplearning}, we introduce some approaches for learning deep representations in an unsupervised~manner.}

\subsection{Event Proposal Generation}
\label{subsec:eventproposalgenerationref}

Event proposal generation is a challenging task because events have no predefined length, ranging from short frame sequences to very long frame sequences with partial or complete overlap.
The general strategy is to define a set of anchors and a deep representation that encodes the video.
Each anchor receives a confidence score from binary classifiers, and the highest-scoring anchors are passed to the captioning module jointly with their associated~representation.

Krishna \textit{et al.}~\citep{krishna:2017}, for example, used a forward sliding window strategy, based on \glspl*{dap}~\citep{escorcia:2016}, with four strides ($1$, $2$, $4$, $8$), to sample video features with different time resolutions and feed them into a \gls*{lstm} unit that encodes and provides past and current contextual information.
On the other hand, Wang \textit{et al.}~\citep{wang:2018} proposed to explore not only past and current context but also the future context to predict and estimate confidence scores.
They adopted a forward and a backward pass on the \gls*{lstm} units and merged the confidence scores using a multiplicative strategy.
They also proposed an attentive fusion approach to compute the hidden representation.
In both works, there are two models, one for each task, trained with an alternate procedure where the proposal module is trained first and then the captioning module is trained while the proposals are~fine-tuned. 

While most works overlooked the intrinsic relationship between the linguistic description and the visual appearance of the events, taking into account only visual features obtained by the C3D model~\citep{tran:2015} pre-trained on the Sports 1-M dataset~\citep{karpathy:2014}, Zhou~\textit{et al.}~\citep{zhou:2018} leveraged the influence of the linguistic description in the proposal module with a vanilla transformer model trained in an end-to-end manner.
Similarly, Iashin and Rahtu~\citep{iashin:2020b} proposed a \gls*{bmt} model using i3D~\citep{carreira:2017} and VGGish~\citep{hershey:2017} features (i.e., visual and audio) to learn video representations conditioned by their linguistic description.
First, the authors trained a captioning model using the ground truth events and sentences.
Then, they used the encoder to feed a multi-headed event proposal module composed of 1D \glspl*{cnn} with different kernel~sizes.

\subsection{Video Captioning}
\label{subsec:videocaptioningref}

Considering the captioning task, most recent methods address this problem in two steps~\citep{vegunopalan:2015,vegunopalan:2015b,donahue:2015}.
In the first step, a neural network encodes the entire video, frame by frame, into a compressed representation given by the hidden state of a \gls*{rnn}.
Then, in the second step, a decoder, usually an \gls*{rnn}, is fed with this representation to learn a probability distribution on a predefined vocabulary, producing a sentence, word-by-word.
More recently, encoder-decoder models based on transformers~\citep{vaswani:2017} have been proposed~\citep{zhou:2018,iashin:2020,iashin:2020b}, however, the best strategy for encoding video information before feeding the encoder remains an open issue.
On the one hand, 2D \gls*{cnn} models can be fed frame by frame, producing long-range feature sequences that are difficult to process using \gls*{rnn} due to the well-known vanishing and exploding gradient problems~\citep{li:2018a}.
\gls*{lstm} and \gls*{gru} combined with soft and hard attention, or even Transformers with self-attention mechanisms, conduct the models to focus on more representative segments. 
These approaches boost performance but do not solve the video representation problem.
On the other hand, when the entire video is fed into a 3D \gls*{cnn} (e.g., as in~\citep{xu:2019}), we come across the problem of information compression.
All semantics are stored in a feature map with a fixed length, and converting this feature map in sentences is difficult because much relevant information can be lost or suppressed --~especially on videos much longer than those used to train the 3D~\gls*{cnn}. 

This problem is more pronounced in captioning than in event proposal generation and has been circumvented by adding modalities such as audio and speech, objects, and action recognition~\citep{pan:2017,gan:2017,iashin:2020,iashin:2020b,chadha:2021}. 
For example, Iashin and Rahtu~\citep{iashin:2020} proposed a framework called \gls*{mdvc}, in which each modality is fed into a separated encoder-decoder transformer and, in the end, their hidden representations are concatenated and fed to a language generator module composed of two dense layers and one softmax~layer. 

Chadha~\textit{et al.}~\citep{chadha:2021} proposed a method to incorporate common-sense reasoning into the \gls*{mdvc} method.
More specifically, they adapted common-sense reasoning from images~\citep{wang:2020b} to videos, thus reaching impressive results in captioning --~especially for the ground truth case.
Although their proposal module uses the new feature to improve the \gls*{bisst} proposal generation method~\citep{wang:2018}, we demonstrate that captioning results can be largely improved by replacing the proposal generation. 

\subsection{Unsupervised Representation Learning}
\label{subsec:unsuperviseddeeplearning}

As discussed earlier, state-of-the-art \gls{dvc} methods employ a combination of multiple modalities of dense representations (e.g., video, audio and speech). In our proposal, we learn a dense representation from visual features in an unsupervised manner by encoding a new semantic information on the videos given by the visual similarities of short clips (clustering) and their co-occurrences (GloVE). This dense representation would replace audio and speech modalities in state-of-the-art~\gls{dvc} methods.

There are a few examples of unsupervised representation learning using clustering in the literature, with remarkable differences from ours. For instance, Xie~\textit{et al.}~\citep{xie:2016} introduced an end-to-end method to learn deep embeddings for cluster analysis. In their approach, a parameterized non-linear mapping is defined to generate a lower-dimensional feature space, where a clustering objective is adopted. Their method was evaluated on image and textual datasets with a few sets of labels (4 and 10) and does not fit our goals.

Another interesting method is DeepCluster, introduced by Caron \textit{et al.}~\citep{caron:2018}. Their approach consists in alternating between clustering of the image descriptors and updating the weights of the convolutional network by predicting the cluster assignments. Similar to Xie~\textit{et al.}~\citep{xie:2016}, they also employ $k$-means but perform a large-scale training of convolutional architectures, incorporating clustering in the architecture and objective.
Finally, Hsu \textit{et al.}~\citep{hsu:2018} also proposed a method to address the problem of effectively grouping visual representations and jointly solve the problem of clustering and representation learning.

The main difference between our proposal and these methods relies on the fact that we employ unsupervised learning to predict soft labels on \textit{short clips} and use these soft labels to generate \textit{visual sentences} in which a \gls{glove} method learns a dense representation for their co-occurrences. Therefore, our features are not optimized to predict a label for the clips but to describe the relationships between the clips.
As mentioned earlier, state-of-the-art ~\gls{dvc} methods take advantage of multi-modal learning. However, it is not easy to provide more modalities for these models for three main reasons: 
(i)~the models will be prone to overfitting due to their increased capacity; 
(ii)~different modalities overfit and generalize at different rates, which requires multiple optimization strategies~\citep{wang:2020};
and (iii)~more preprocessing is necessary to produce the features. 
We provide a relevant contribution by extracting more video information using only visual features without human annotations.
Our method is an improved bag-of-word approach widely used in computer vision. However, it has not yet been applied to dense video captioning to the best of our knowledge. Additionally, this type of information (i.e., co-occurrence similarity) is not easily learned by deep learning techniques, especially in an unsupervised way, justifying our choice for the combination of $k$-means and~\gls{glove}.

%% file: 3-methodology.tex
\section{Methodology}
\label{sec:methodology}

In this work, we propose a dense video captioning system that leverages unsupervised semantic information and is trained in two steps.
In the first step, a temporal event proposal module is responsible for generating central points, event lengths, and confidence scores, predicting whether an event is contained in that location.
This proposal generator is trained by adopting the architecture and procedures from Iashin and Rahtu~\citep{iashin:2020b}, which are described in this section.
Nevertheless, we replace the audio signal with the proposed semantic descriptor.
Figure~\ref{fig:architecture} shows the main elements of this step: a bi-modal transformer, a proposal generator, and a language~generator.

\input{figures/architecture/fig_architecture}

\glsreset{bmt}
In the second step, we employ the vanilla transformer used by Iashin and Rahtu~\citep{iashin:2020}, replacing the \gls*{bisst} proposal module with the \gls*{bmt} proposal module due to their state-of-the-art performance in event proposal generation.
The main elements in this step are a vanilla transformer and a language generator.
In both of them, we employ the proposed semantic descriptor, shown in Figure~\ref{fig:architecture}, for the element co-occurrence estimation.

Bi-modal and vanilla transformers are composed of encoder and decoder layers. As vanilla is the base for the construction of bi-modal, we first explain how the captioning module works and then how the proposal generator~works.

\subsection{Co-Occurrence Similarity Estimation Module}
\label{sec:cooccurence}

Let $D_{Tr}=\{V_{Tr_{1}},..., V_{Tr|D_{Tr}|}\}$ and $D_{Te}=\{V_{Te_{1}},..., V_{Te|D_{Te}|}\}$ be the training and testing datasets, respectively, composed of videos with long duration (e.g., $1$--$2$~min) and with more than one event per video.
We first take all videos from $D_{Tr}$ and split each one into short clips with $f$ frames each.
Then, we sample all these short clips and extract features using the i3D model~\citep{carreira:2017}.
As a result, a set of features $X=\{x_{1},x_{2},...,x_{l}\}$, where $l=\lfloor n_f/f \rfloor$ and $n_f$ is the number of frames of a given video, with $ x \in \mathbb{R}^{1024}$ is produced per video.
Next, a mini-batch $k$-means algorithm~\citep{sculley:2010} is trained to minimize the Euclidean distance
\begin{equation}
    \label{eq:clusteringobjective}
    \min{\sum_{x \in X} ||Ecd(C,x)-x||^2} \, ,
\end{equation}
\noindent where $Ecd(C,x)$ stands for the nearest cluster center $c \in C$ to $x$ and $|C|$ corresponds to our codebook size (e.g., $1{,}500$~clusters).

Once we have trained the clustering model, a video can be processed by first splitting it into clips of $f$ frames and then extracting the i3D features (using only the RGB stream) from these clips assigning each one of them to a cluster.
These sequences of labeled clusters build a storytelling, and we can learn information about their co-occurrence properties, similarly to the dense representation from the GloVe method~\citep{pennington:2014}.

We compute a matrix of co-occurrence counts, denoted by $Z$, whose entries $Z_{ij}$ tabulate the number of times the cluster $j$ occurs in the context $S$ (an arbitrary sliding window) of cluster $i$. 

Let $Z_{i}=\sum_{k}Z_{ik}$ be the number of times any cluster appears in the context of cluster $i$, we define the co-occurrence probability as
\begin{equation}
    \label{eq:co-occurrence}
    P_{ij} = P(j|i) = \frac{Z_{ij}}{Z_{i}}.
\end{equation}

Pennington {\em et al.}~\citep{pennington:2014} showed that the vector learning should be with ratios of co-occurrence probabilities rather than with the probabilities themselves, as this choice forces a greater difference in values between clusters that occur close frequently compared to infrequent cases.
This ratio can be computed considering three clusters $i$, $j$ and $k$ with ($P_{ik}/P_{jk}$) and the model takes the general form given~by
\begin{equation}
    \label{eq:modelformulation}
    F(w_{i},w_{j},\tilde{w}_{k}) = \frac{P_{ik}}{P_{jk}} \, ,
\end{equation}
\noindent where $w \in \mathbb{R}^{128}$ are cluster vectors and $\tilde{w} \in \mathbb{R}^{128}$
are separate context cluster vectors.
Our model is a weighted least square regression trained with a cost function given by
\begin{equation}
    \label{eq:costfunction}
    J = f(Z_{ij})\sum_{i,j=1}^{|C|}{(w_{i}^{T}\tilde{w}_{j} + b_{i} + \tilde{b}_{j} - \log{Z_{ij}})^{2}} \, ,
\end{equation}
\noindent where $|C|$ is the size of the vocabulary (i.e., $1{,}000$ clusters), $b_{i}$ and $\tilde{b}_{j}$ are bias vectors and $f$ is a weighted function defined~as
\begin{equation}
    \label{eq:weightedfunction}
    f(t) =
    \left\{
    	\begin{array}{ll}
    		(t/t_{\max})^{\alpha}  & \mbox{if } t < t_{\max} \\
    		1 & \mbox{otherwise } 
    	\end{array}
    \right. ,
\end{equation}
\noindent where $t_{\max}=100$ and $\alpha=3/4$. 
More details and a complete mathematical description are provided in~\citep{pennington:2014}.
For our purposes, we adopt $w$ as our semantic descriptor, represented as $\textit{Sm}$ in the remainder of this~work.

\subsection{Video Captioning Module}

Given a video $V$, the video captioning module takes a set of $n_c$ visual features $V_{f}=\{v_{f_1},...,v_{f_{n_c}}\}$, one per each clip, and a set of $m$ words $Y=\{y_1,...,y_m\}$ to estimate the conditional probability of an output sequence given an input~sequence. 

We encode $v_{f_c}$, where $1 \leq c \leq n_c$, as a concatenation of features defined as
\begin{equation}
\label{eq:visualfeature}
v_{f_c}= [V_{E}(v_{c}), \textit{Sm}(v_{c})] \, ,
\end{equation}
\noindent where $V_{E}(\cdot)$ yields a deep representation given by an off-the-shelf neural network (e.g., i3D~\citep{carreira:2017} with RGB or RGB + \gls*{of} streams), $\textit{Sm}(\cdot)$ produces our co-occurrence similarity representation (see Section~\ref{sec:cooccurence}), $[\text{ }]$~is a concatenation operator, and $v_c$ is the $c$-th short clip for the video~$V$.

The video features are fed to the original transformer model~\citep{vaswani:2017}, composed of several layers (as shown in Figure~\ref{fig:architecture}), in which an encoder maps a sequence of visual features to a continuous representation that is used by a decoder to generate a sequence of symbols~$Y$. 

First, the visual embedding of each video is computed using Equation~\ref{eq:visualfeature} and feeds all at once.
Then, to provide information on the position of each feature we employ the same encoding method used by Vaswani~{\em et al.}~\citep{vaswani:2017}, a position-wise layer computes the position with sine and cosine at different frequencies as~follows
\begin{eqnarray}
    \label{eq:poswise}
    \textit{PE}_{(pos, 2i)} & = & \sin{(pos/10000^{2i/d_{model}})}, \\
    \textit{PE}_{(pos, 2i+1)} & = & \cos{(pos/10000^{2i/d_{model}})}, \nonumber
\end{eqnarray}
\noindent where $pos$ is the position of the visual feature in the input sequence, $0 \leq i < d_{model}$ and $d_{model}$ is a parameter defining the internal embedding dimension in the transformer.

In the encoder, these representations are passed through a multi-head attention layer. The attention used is the scaled dot-product and is defined in terms of queries ($Q$), keys ($K$), and values ($V$) as
\begin{equation}
    \label{eq:attention}
    \textit{Att}(Q,K,V)=\textit{softmax}(\frac{QK^{T}}{\sqrt{d_{k}}})V.
\end{equation}

The multi-head attention layer is defined by the concatenation of several heads ($1$ to $h$) of attention applied to the input projections as
\begin{equation}
    \label{eq:multihead}
    \textit{MHAtt}(Q,K,V)=[head_{1},..., head_{h}]W^{0} \, ,
\end{equation}
\noindent where $\textit{head}_{i}=\textit{Att}(QW_{i}^{Q},KW_{i}^{K},VW_{i}^{V})$ and $[\text{ }]$ is a concatenation operator.

Once we compute self-attention, $Q = K = V = V_{f}^{\textit{PE}}$, 
which results in 
\begin{equation}
\begin{split}
V_{f}^{self-att}=[\textit{Att}(V_{f}^{\textit{PE}}W_{i}^{V_{f}^{\textit{PE}}},V_{f}^{\textit{PE}}W_{i}^{V_{f}^{\textit{PE}}},V_{f}^{\textit{PE}}W_{i}^{V_{f}^{\textit{PE}}}), \\
...,\textit{Att}(V_{f}^{\textit{PE}}W_{h}^{V_{f}^{\textit{PE}}},V_{f}^{\textit{PE}}W_{h}^{V_{f}^{\textit{PE}}},V_{f}^{\textit{PE}}W_{h}^{V_{f}^{\textit{PE}}})].    
\end{split}
\end{equation}

At the end of each encoder layer, a fully connected feed-forward network $\textit{FFN}(\cdot)$ is applied to each position separately and identically.
It consists of two linear transformations with a ReLU activation and is defined as
\begin{equation}
    \label{eq:ffn}
    \begin{split}
    \textit{FFN}(u) = \max(0, uW_{1}+b_{1})W_{2}+b_{2} 
    \end{split} \; ,
\end{equation}
\noindent resulting in $V_{f}^{\textit{FFN}}$ that is used in the decoder layer.

The decoder layer receives words and feeds an embedding layer $\textit{E}(\cdot)$, computing a position with Equation~\ref{eq:poswise} resulting in $W^{\textit{PE}}$.
Then, this representation is fed to the multi-head self-attention layer (see Equation~\ref{eq:multihead}), resulting in $W^{self-att}$.
At this moment, the visual encoding provided by encoder layers feeds a multi-head attention layer as
\begin{equation}
\label{eq:multiheaddecoder}
W^{VisAtt} = \textit{MHAtt}(W^{self-att}, V_{f}^{\textit{FFN}},V_{f}^{\textit{FFN}}).
\end{equation}

Finally, $W^{VisAtt}$ feeds an $\textit{FFN}(\cdot)$ and, then, a generator $G(\cdot)$ composed of a fully connected layer and a softmax layer is responsible for learning the predictions over the vocabulary distribution~probability.

\subsection{Event Proposal Module}
\label{sec:bmt_model}

The event proposal module uses the bi-modal transformer.
Considering the encoder, this transformer has two differences from the vanilla encoder.
It takes two streams, visual $V_{f}$ and semantic $\textit{Sm}$, separately, and it has three sub-layers in the encoder: self-attention (Equation~\ref{eq:attention}), producing $V_{f}^{self-att}$ and $\textit{Sm}^{self-att}$; 
bi-modal attention, i.e.,
\begin{eqnarray}
V_{f}^{\textit{Sm}-att} & = &\textit{MHAtt}(V_{f}^{self},\textit{Sm}^{self}, \textit{Sm}^{self}),
\label{eq:visualsemanticattended} \\
\textit{Sm}^{Vis-att} & = &\textit{MHAtt}(\textit{Sm}^{self}, V_{f}^{self},V_{f}^{self}),
\label{eq:semanticvisualattended}
\end{eqnarray}

\noindent and a fully connected layer $\textit{FFN}(\cdot)$ for each modality attention, producing $V_{{\textit{Sm}-att}}^{\textit{FNN}}$ and $\textit{Sm}_{v-att}^{\textit{FNN}}$ used in the bi-modal attention unit on the decoder and in the multi-headed proposal generator. 

In the bi-modal decoder, the differences to the vanilla decoder are the bi-modal attention and bridge layers.
First, a $W^{self-att}$ is obtained with Equation~\ref{eq:multihead}.
Afterward, the bi-modal attention is computed as
\begin{equation}
\label{eq:wordsmatt}
W^{\textit{Sm}-att}=\textit{MHAtt}(W^{self-att}, \textit{Sm}_{v-att}^{\textit{FNN}},\textit{Sm}_{v-att}^{\textit{FNN}}) \, ,
\end{equation}
\noindent and
\begin{equation}
\label{eq:wordvisatt}
W^{V-att}=\textit{MHAtt}(W^{self-att}, V_{{\textit{Sm}-att}}^{\textit{FNN}},V_{{\textit{Sm}-att}}^{\textit{FNN}}).
\end{equation}

The bridge is a fully connected layer on the concatenated output of bi-modal attention mechanisms, given as
\begin{equation}
\label{eq:z}
W^{\text{FFN}}=\text{FFN}([W^{\textit{Sm}-att}, W^{V-att}]).
\end{equation}

The output of the bridge is passed through another $\textit{FFN}$ and then to the generator $G(\cdot)$.
This means that the encoder parameters are learned in the captioning task, improving the visual features by conditioning them to the vocabulary. 

More specifically, we focus on the $\textit{Sm}_{v-att}^{\textit{FNN}}$ and $V_{{\textit{Sm}-att}}^{\textit{FNN}}$ outputs. 
The proposal heads take these embeddings and make predictions for each modality individually, forming a pool of cross-modal predictions.
The process begins with defining a $\Psi$ set of anchors with a central location and a prior length.
A fully connected model with three 1D convolutional layers (with kernels $k_{1}=\text{arbitrary}$, $k_{2}=k_{3}=1$) predicts the value for the length and confidence score for each anchor.
Then, these predictions are grouped and sorted by their confidence levels, preserving the proportionality between the source modalities.
The process of selecting a $\Psi$ set of anchors follows the common approach of learning a $k$-means clustering model by grouping similar lengths using the ground-truth~annotations~\citep{krishna:2017,wang:2018,iashin:2020, chadha:2021}.

\subsection{Training Procedure}
\label{subsec:trainingprocedure}

The first stage is the training of the semantic descriptor. We split each video from the training set into clips with $f=64$ frames and compute the i3D representation with only the RGB stream for each clip. 
Then, a mini-batch $k$-means learns a codebook with $|C|=1500$ visual words in a procedure with 5 epochs.
Once we have learned the clustering model, the semantic embedding is trained using a sliding window $S=5$, corresponding to $\approx 10$ seconds and cluster embedding vectors with 128 dimensions. 
The training occurs up to 1500 iterations, with early stopping applied after 100 iterations. 
The Adagrad optimizer~\citep{duchi:2011} with learning rate $lr=0.05$ is used.

The second stage is the training of the bi-modal encoder conditioned by the vocabulary. 
Thus, a captioning model is learned, using teaching forcing in which the target word is used as next input, instead of the predicted word, optimizing the \textit{KL-divergence loss}, applying Label Smoothing~\citep{szegedy:2016} to make the model less confident over frequent words, and applying masking to prevent the model from attending on the next positions on the ground-truth sentences.

The model is learned up to 60 epochs with early stopping to monitor the METEOR score~\citep{banerjee:2005}, using the Adam optimizer~\citep{diederik:2015} with $\beta_{1}=0.9$, $\beta_{2}=0.999$, $lr=5.10^{-5}$ and $\epsilon=1.10^{-8}$. 
These procedures are also adopted in the final captioning training (i.e., using the vanilla~transformer).

The bi-modal encoder is used to learn the multi-head proposal module with \gls*{mse} for localization losses and cross-entropy for confidence losses.
Then, we learn the final captioning model feeding the vanilla transformer~\citep{vaswani:2017} with ground-truth proposals and sentences. Thus, we predict the sentences to assess the~performance.

\minor{All experiments were conducted on a computer equipped with an AMD Ryzen 7 2700X 3.7GHz CPU, 64 GB of RAM, and an NVIDIA Titan Xp GPU (12 GB).
The experiments were performed using the Ubuntu operating~system.}

%% file: figures/architecture/fig_architecture.tex
\begin{figure*}[!htb]
    \captionsetup{captionskip=0.25pt}
    \centering
    \resizebox{0.975\linewidth}{!}{
	\subfloat[][]{
        \includegraphics[height=37ex]{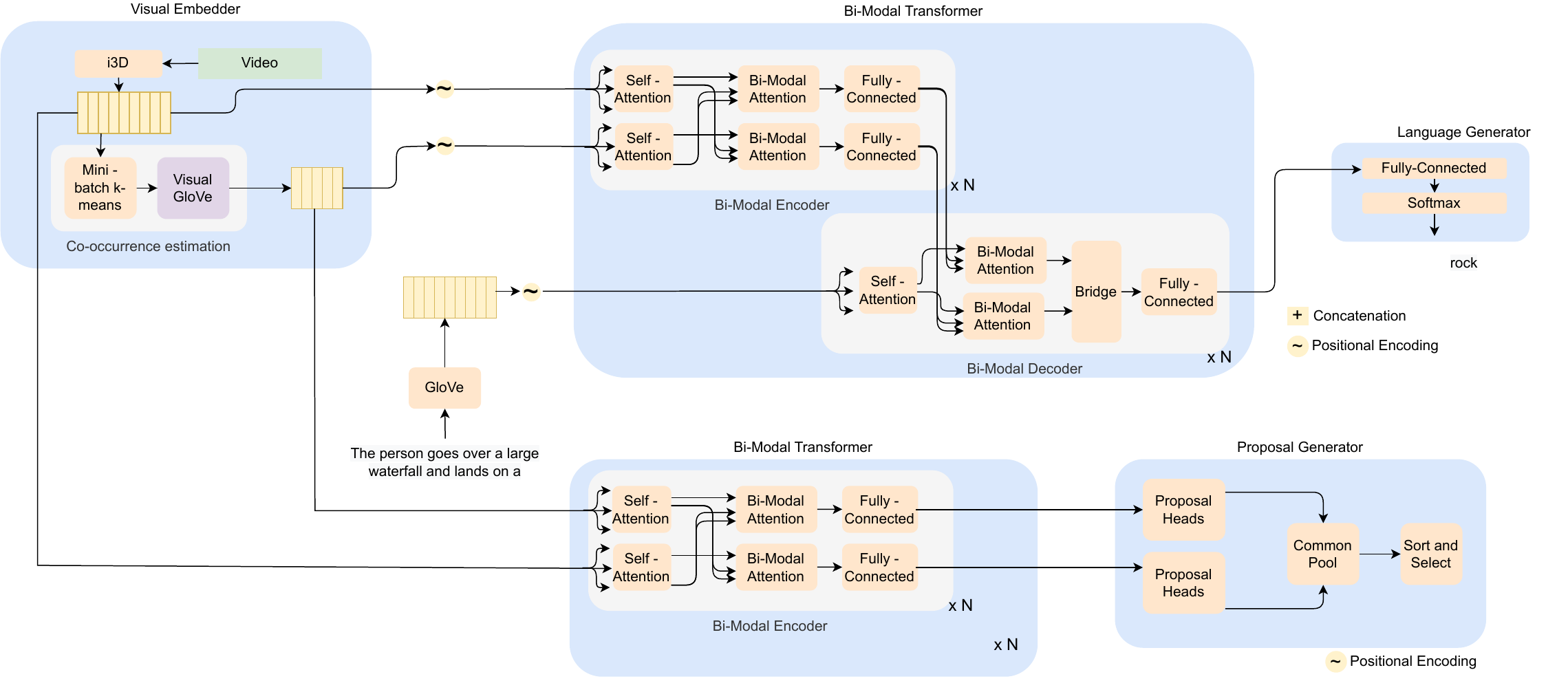} 
        }
    }
    
    \vspace{4mm}
    
    \resizebox{0.975\linewidth}{!}{
	\subfloat[][]{
        \includegraphics[height=17ex]{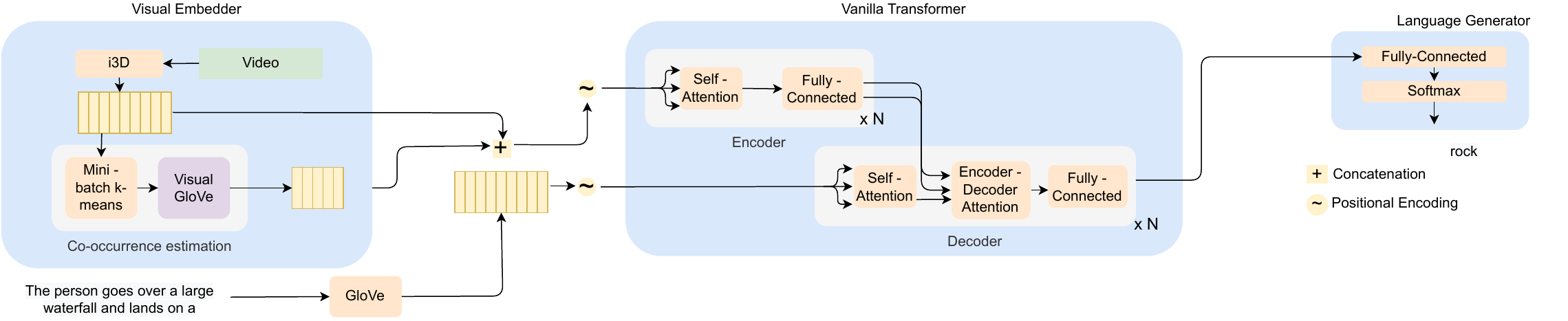} 
        }
    }

    \vspace{-2mm}
    
    \caption{\minor{Overview of the proposed method. (a)~describes the event proposal phase, which consists of two stages. In the first stage, a bi-modal transformer is employed in a captioning task, where visual and semantic co-occurrence-based features are used to learn the encoder parameters conditioned by language. Then, in the second stage, these encoder parameters are used to predict temporal event proposals. In the second stage~(b), these proposals are used to generate captions using a vanilla transformer and a language generator trained with ground-truth events and~sentences.}} 

    \label{fig:architecture}
\end{figure*}

%% file: 4-protocol.tex
\section{Dataset and Evaluation Metrics}

All experiments were performed on the ActivityNet Captions dataset~\citep{krishna:2017}, which is a large-scale dataset with temporal segments annotated and described in the proportion of one sentence for each segment. ActivityNet Captions was selected because it is a challenging open-domain dataset used as a default evaluation by all reference works. The dataset contains $20{,}000$ videos divided into training/validation/test subsets with $50$/$25$/$25$\% videos, respectively, and $3.65$ events per video on average.
As the annotations of the test set are not public, we used the validation set for testing, as in previous works~\citep{wang:2018,chadha:2021,iashin:2020,iashin:2020b}.

The validation set was annotated twice (val1 and val2), and we consider the average for each evaluation metric on each validation split.
The captioning task was evaluated using the BLEU@1-4~\citep{papineni:2002}, METEOR~\citep{banerjee:2005}, ROUGE${_L}$~\citep{lin:2004} and CIDEr-D~\citep{vedantam:2015} metrics computed with the evaluation script provided by Krishna~{\em et al.}~\citep{krishna:2017}, whereas event proposal was evaluated with Precision, Recall and F1-score (i.e., the harmonic mean of precision and~recall).

%% file: 5-results.tex
\section{Results}
\label{sec:results}

\minor{This section discusses our results on event proposal generation and video captioning.
We provide a comparison with \gls*{sota} methods, a qualitative analysis, and ablation studies.}

\subsection{\minor{Results for Event Proposals}}

As described in Section~\ref{subsec:trainingprocedure}, we explored the captioning training to learn the parameters of the bi-modal encoder and then used this encoder to predict the proposals in the bi-modal proposal generator module.
Afterward, these proposals were employed in a vanilla transformer captioning model.

Table~\ref{tab:bmt_cap_comparison} shows the \gls*{bmt} performance on video captioning.
We highlighted as baselines the results from Iashin and Rahtu~\citep{iashin:2020b} with only visual features (i.e., using a vanilla transformer) and with bi-modal features (i.e., using visual and audio features denoted by $\text{BMT}$).
Additionally, we included the performance from Iashin and Rahtu~\citep{iashin:2020} with visual, audio, and speech modalities and employing \gls*{bisst} as the event proposal module.
Lastly, we investigated how \gls*{bmt} captioning performs with $\textit{RGB+Sm}$ and with $\textit{V+Sm}$ (i.e., $\textit{RGB~+~OptFlow~+~Sm}$).

\input{tables/bmt_results}

Our results with $\textit{V+Sm}$ presented superior performance compared to the Visual performance from~\citep{iashin:2020b} considering all metrics and proposals schemes (GT and Learned).
Comparing $\text{BMT}$ with $\text{BMT}_\textit{V+Sm}$, we observed a slightly lower performance using $\textit{Sm}$ instead $\textit{A}$ (audio) considering all scores.

However, the proposed model was still capable of learning a high-quality encoder, as evidenced by the performance levels achieved on proposal generation (see Table~\ref{tab:event-proposal-generation}).
We reached competitive results in terms of $\text{F1}$-score and Precision compared to the original $\text{BMT}$ in the $\textit{V+Sm}$ scenario.
This slight difference in F1-score supports the adoption of only visual features for event proposal generation due to the fewer preprocessing requirements than BMT.
Considering the performance on $\textit{RGB+Sm}$ configuration (i.e., even less preprocessing), we outperformed the popular \gls*{bisst} method while achieving competitive performance with \gls*{bmt}.
Masked transformer~\citep{zhou:2018}, which is a method that explores only visual features and linguistic information to learn temporal proposals, is outperformed by our approach in $11.8$\% [i.e., 59.60/53.31] in terms of F1-score.

\input{tables/event_proposals_results}

\subsection{\minor{Results for the Video Captioning Stage}}

In Table~\ref{tab:stateoftheart}, we show a comparison between our results \minor{using the Vanilla Transformer, fed by the concatenation of visual and semantic descriptors,} and the results obtained by \gls*{sota} methods.
As can be seen, there are methods based only on visual features and methods based on multi-modal features (see  column \textit{VF}).
As the videos from ActivityNet captions must be downloaded from YouTube, several videos have become unavailable since the original dataset was published.
Hence, we used $91$\% of the dataset (this information is presented in column $\text{FD}$, where a ``\cmark'' means that $100$\% of the videos were available at the time of the experiments).
As we have a reduced set of videos for evaluation, the validation sets were filtered to contain only the videos downloaded.
As demonstrated in~\citep{iashin:2020b}, this procedure enables a fair comparison because the \textit{\gls*{sota} methods reached almost unchanged results} when evaluated using these filtered validation sets. 
However, not considering this procedure is unfair, because the model is forced to propose events and generate captions for unseen videos, reducing performance.
Finally, some works adopted a direct optimization of the METEOR score with reinforcement learning techniques (see column~\text{RL}).
We also listed the performance without these techniques since, as shown in Table~\ref{tab:stateoftheart} for DVC~\citep{li:2018}, these techniques boosted the METEOR score without a proportional boost in BLEU, which may not corresponds to an actual improvement in the captioning~quality.

\input{tables/sota}

Considering only the single modality scenario, without $\textit{RL}$, our model outperforms all other methods in learned proposals and has a slightly lower performance on BLEU@3-4 than the Masked Transformer for GT proposals.
Compared to the multi-modal methods, our performance on ground truth is lower than the \gls*{mdvc} and iPerceive methods. 
However, we remark that the performance on GT proposals is an indicator of how good the captions are when the event is perfectly delimited.
As can be seen in Table~\ref{tab:event-proposal-generation}, this ideal scenario is far from being the case, and the most significant performance to consider is in the learned proposals scenario, where our results are particularly~noteworthy.

Finally, we highlight the results of the TSP method~\citep{alwassel:2020} compared to ours.
This method includes an improved visual descriptor for temporal event localization that combines local features optimized by accuracy on \gls*{tac} and global features given by pooling local predictions. 
The authors employed the R(2+1)D architecture~\citep{tran:2018} fine-tuned on the ActivityNet v1.3 dataset~\citep{heilbron:2015}.
They adopted the \gls*{bmt} model for captioning, and a critical procedure for the success of video captioning was the fine-tuning on ActivityNet with trimmed action annotations (METEOR of $8.75$ with fine-tuning and $8.42$ without~\citep{alwassel:2020}).
Lastly, their model also considers the audio signal.
Thus, it is noteworthy that our model reaches a comparable performance on METEOR without audio and action~annotations.

\subsection{\minor{Qualitative Analysis}}

\begin{figure}
     \centering
     \captionsetup[subfigure]{captionskip=0pt} 
     
     \resizebox{0.85\linewidth}{!}{
     \subfloat[][]{\includegraphics[height=24ex]{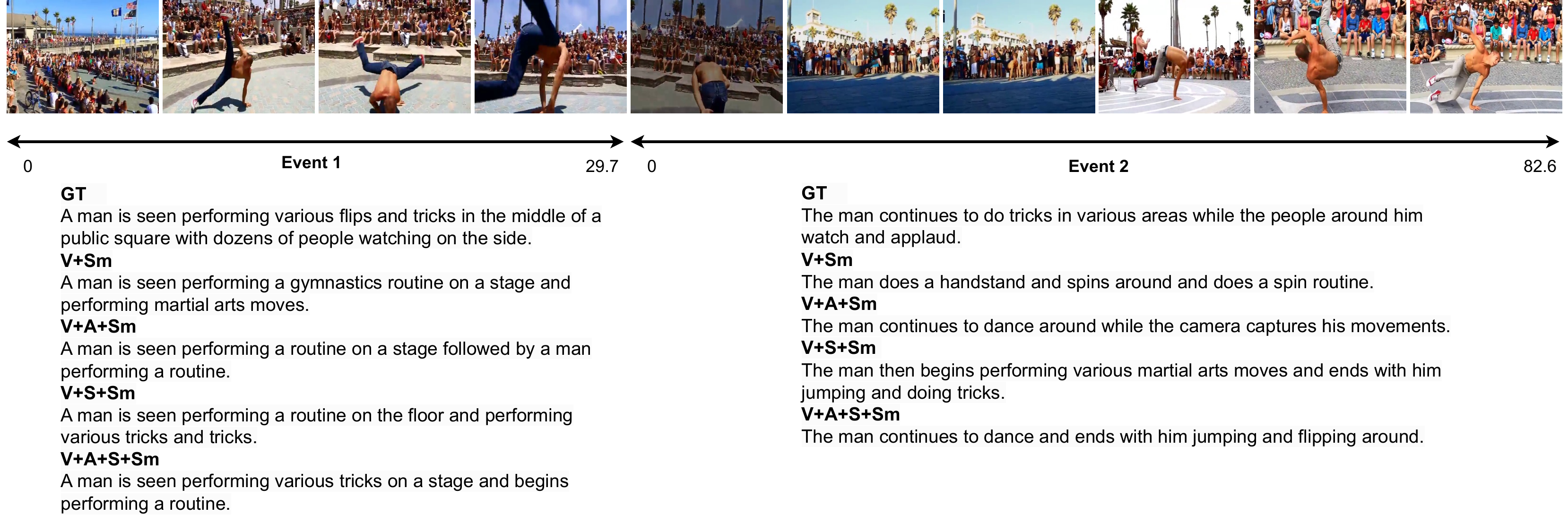}\label{fig:q_a}}
     }

     \vspace{4mm}
     \resizebox{0.85\linewidth}{!}{
     \subfloat[][]{\includegraphics[height=22ex]{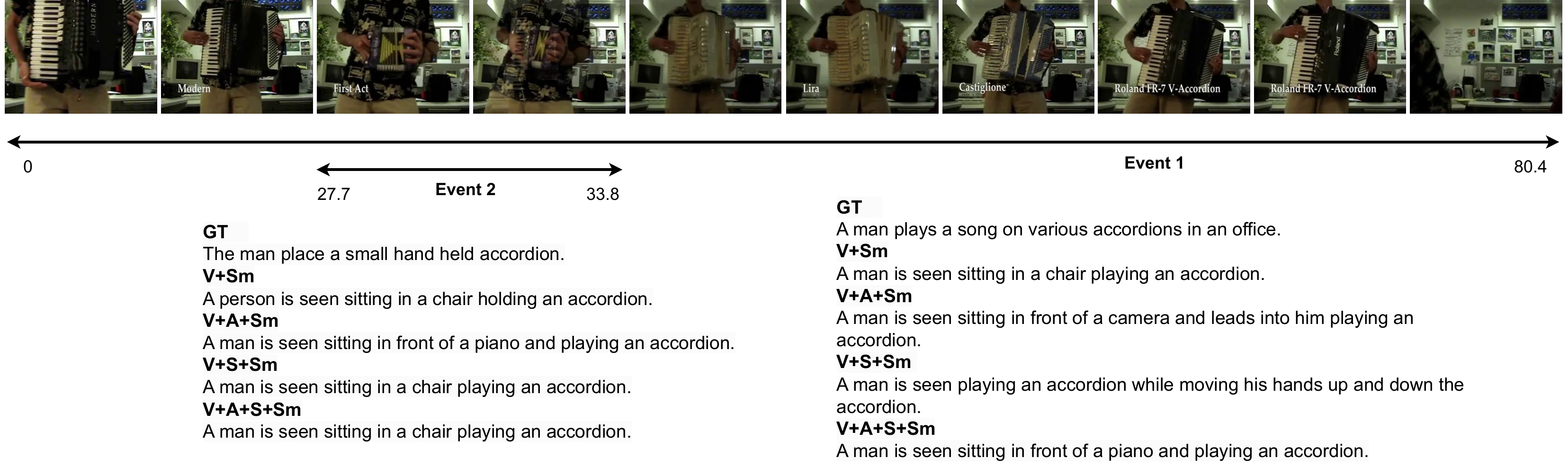}\label{fig:q_b}}
     }

     \vspace{4mm}
     \resizebox{0.85\linewidth}{!}{
     \subfloat[][]{\includegraphics[height=23ex]{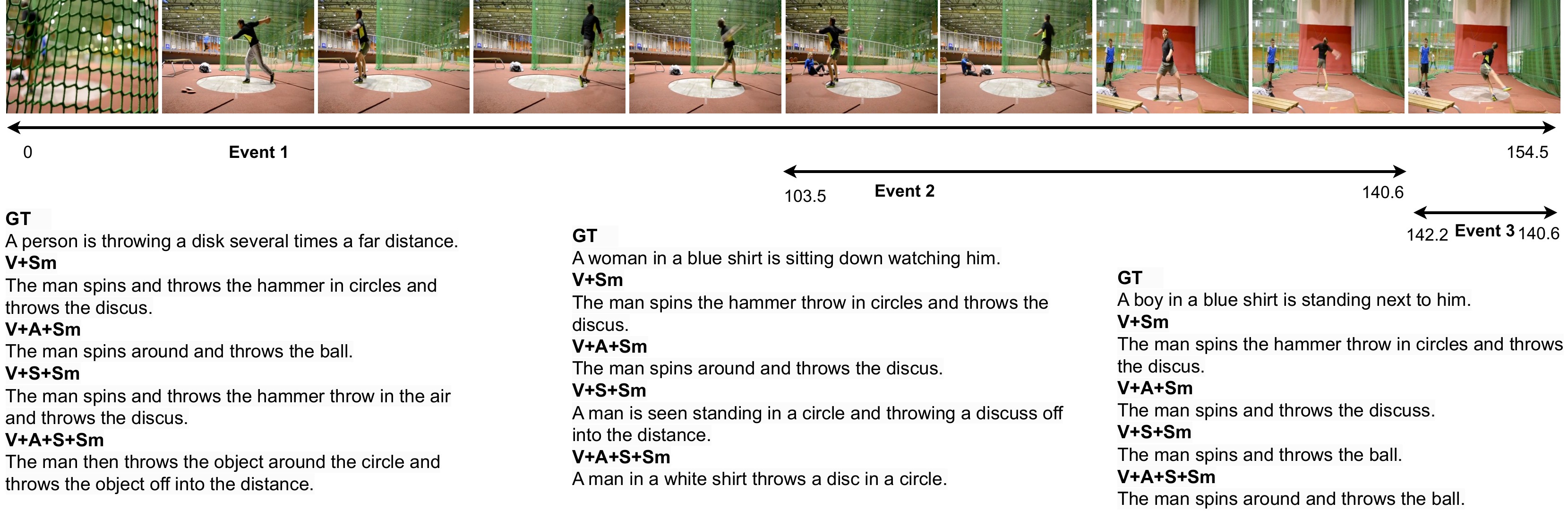}\label{fig:q_c}}
     }

     \vspace{4mm}
     \resizebox{0.85\linewidth}{!}{
     \subfloat[][]{\includegraphics[height=24ex]{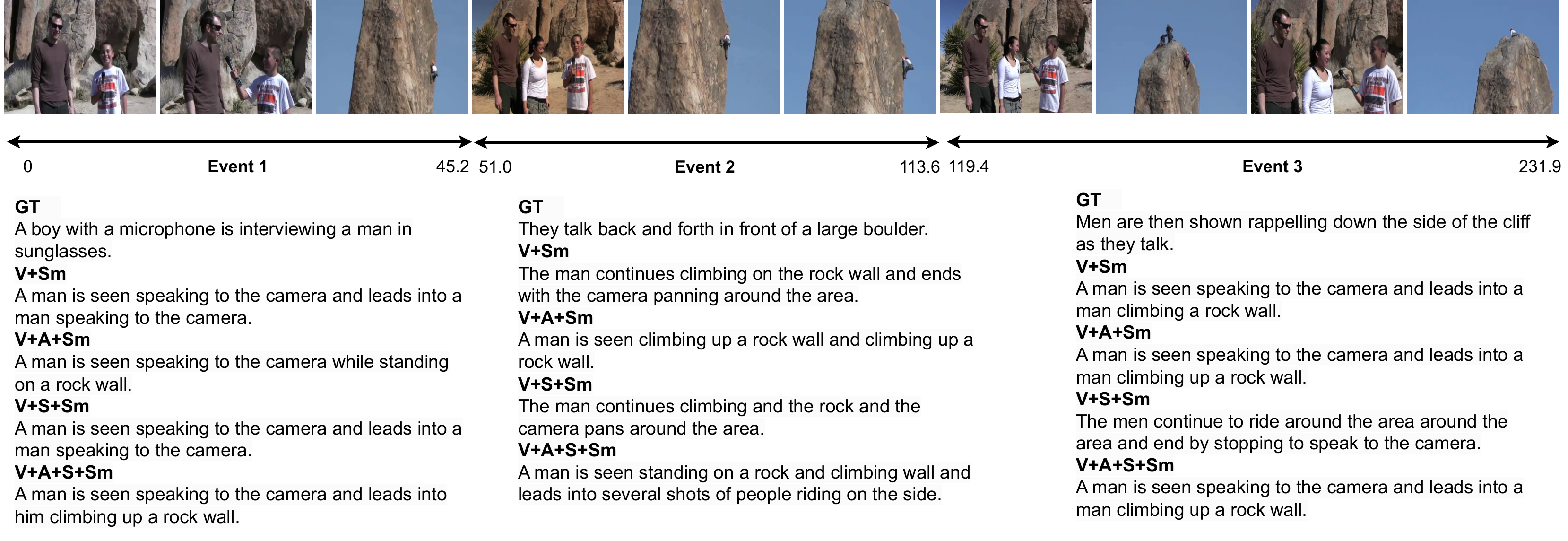}\label{fig:q_d}}
     }

     \vspace{-1.5mm}
     
     \caption{Qualitative comparison on the results including our semantic descriptor in the \gls*{mdvc} method. We show results for event proposals captioning considering the highest Intersection over Union for each ground truth event proposal in the following videos (a) svWiQtzgtOc, (b) P4PQ5tC3gX8, (c) BhAQhPasmhU, and (d) 045Tkq12H\_c.}
     \label{fig:qualitativeanalysis}
\end{figure}

\minor{In Figure~\ref{fig:qualitativeanalysis}, we show a qualitative analysis of dense captioning considering our semantic descriptor in combination with audio and speech.
The model we propose is denoted as configuration \textit{V+Sm}.
In Section~\ref{subsec:mdvccombinedvisglove}, we provide the quantitative results of these combinations.
We selected four videos in different contexts to evaluate the ability to describe general events. The selected videos are svWiQtzgtOc, P4PQ5tC3gX8, BhAQhPasmhU, and 045Tkq12H\_c.}

\minor{Upon evaluating the examples in Figure~\ref{fig:q_a}, it can be observed that Event~1 posed a challenge for models \textit{V+Sm} and \textit{V+A+Sm}, as they confused street dance with a gymnastics routine.
Indeed, the movement is very similar to the one performed on the pommel horse.
In Event~2, only \textit{V+A+Sm} and \textit{V+A+S+Sm} managed to correctly identify it as a dance.
\textit{V+Sm} described the move exactly, but even so, it would be a description with low scores B@3-4 or M.
When analyzing the sentences in Figure~\ref{fig:q_b}, for Event~1, none of the models were able to describe the various accordions throughout the video, although all reported that there is an accordion.
This highlights the challenges in recognizing information with long-range dependencies.
As for Figure~\ref{fig:q_b}, Event~2, the models failed to predict the person~standing.}

\minor{In Figure~\ref{fig:q_c}, Event 1, only V+S+Sm correctly described that the object being thrown was a discus, while all models correctly described the spinning and throwing motion.
In Event 2, no model described the woman in blue.
This indicates that the models tend to prioritize the foreground objects.
Nevertheless, a less observant person would likely overlook the woman as well.
Similarly, in Event 3, none of the models described the boy.
Finally, in Figure~\ref{fig:q_d}, the models correctly identify what is happening, but they generated sentences with syntactic problems, notably the repetition of fragments. 
In general, there was no significant improvement in sentence quality when incorporating additional modalities such as \textit{A} or \textit{S}, reinforcing the potential of our semantic descriptor combined with the \textit{V}~modality.}

\input{5.4-ablations}

%% file: tables/bmt_results.tex
\begin{table}[!htb]
\centering
\caption{Captioning performance comparison of the BMT and Transformer methods with different features in the same validation sets.
For each metric, the top $2$ results are highlighted in bold.}
\vspace{-2mm}
\footnotesize
\resizebox{.50\linewidth}{!}{
\begin{tabular}{lcccccc}
\toprule
& \multicolumn{3}{c}{GT Proposals} & \multicolumn{3}{c}{Learned Proposals}      \\
& B@3        & B@4        & M           & B@3        & B@4        & M           \\ 
\midrule

Visual~\citep{iashin:2020b}        &
3.77       & 1.66       & 10.29       & 2.85       & 1.30       & 7.47          \\

$\text{BMT}$~\citep{iashin:2020b}  &
\tbf{4.62} & \tbf{1.99} & \tbf{10.89} & \tbf{3.84} & \tbf{1.88} & \textbf{8.44} \\

MDVC$_{\textit{Bi-SST}}$           &
\tbf{4.52} & \tbf{1.98} & \tbf{11.07} & 2.53       & 1.01       & 7.46          \\

\midrule \midrule

BMT$_{\textit{RGB+Sm}}$            &
4.12       & 1.72       & 10.32       & 3.62       & 1.74       & 8.03          \\ 

BMT$_{\textit{V+Sm}}$              &
4.32       & 1.85       & 10.55       & \tbf{3.68} & \tbf{1.81} & \textbf{8.26} \\  

\bottomrule
\end{tabular}
}
\label{tab:bmt_cap_comparison}
\end{table}

%% file: tables/event_proposals_results.tex
\begin{table}[!htb]
\centering
\caption{Comparison with state-of-the-art proposal generation. Results are reported on the validation sets using Precision, Recall and F1-score and are taken for $100$ proposals per video ratio.
For each metric, the top $2$ results are highlighted in bold.}
\vspace{-2mm}
\footnotesize
\begin{tabular}{lcccc}
\toprule
                                    & FD       &   Prec.        &  Rec.           &  F1         \\ 
\midrule
MFT~\citep{xiong:2018}              & \cmark   & 51.41          & 24.31           & 33.01       \\
BiSST~\citep{wang:2018}             & \cmark   & 44.80          & 57.60           & 50.40       \\
Masked Transf.~\citep{zhou:2018}    & \cmark   & 38.57          & \tbf{86.33}     & 53.31       \\
SDVC~\citep{mun:2019}               & \cmark   & \tbf{57.57}    & 55.58           & 56.56       \\
BMT$_{V+A}$~\citep{iashin:2020b}    & \xmark   & 48.23          & \tbf{80.31}     & \tbf{60.27} \\
PDVC~\citep{wang:2021e2e}           & \xmark   & \tbf{58.07}    & 55.42           & 56.71       \\
Ours$_{RGB+Sm}$                     & \xmark   & 47.27          & 78.71           & 59.07       \\
Ours$_{V+Sm}$                       & \xmark   & 48.11          & 78.31           & \tbf{59.60} \\
\bottomrule
\end{tabular}
\label{tab:event-proposal-generation}
\end{table}

%% file: tables/sota.tex
\begin{table}[!htb]
\centering
\caption{Comparison with other methods on ActivityNet Captions (validation set). 
VF~=~Visual features only; RL~=~Reinforcement Learning -- reward maximization (METEOR); FD~=~Full dataset was~available. The top $2$ results are highlighted in bold.}
\footnotesize
\vspace{-2mm}
\setlength{\tabcolsep}{3pt}
\resizebox{.60\linewidth}{!}{
\begin{tabular}{lccccccccc}
\toprule
& \multirow{2}{*}{VF} & \multirow{2}{*}{RL}     & \multirow{2}{*}{FD} & \multicolumn{3}{c}{GT Proposals} & \multicolumn{3}{c}{Learned Proposals} \\
&        &        &        & B@3   & B@4    & M     & B@3   & B@4   &  M   \\ 
\midrule
DVC \citep{li:2018}              
& \cmark & \cmark & \cmark & 4.55  & 1.62   & 10.33 & 2.27  & 0.73  & 6.93 \\

SDVC~\citep{mun:2019}            
& \cmark & \cmark & \cmark & 4.41  & 1.28   & 13.07 & 2.94  & 0.93  & \textbf{8.82} \\ 

\minor{GVL~\citep{wang2023learning}} 
& \cmark & \cmark & \xmark & $-$   & $-$    & $-$   & $-$   & 1.11 & 10.03     \\

\midrule

Dense Cap~\citep{krishna:2017}   
& \cmark & \xmark & \cmark & 4.09  & 1.60   & \phantom{0}8.88  & 1.90  & 0.71  & 5.69 \\

DVC \citep{li:2018}              
& \cmark & \xmark & \cmark & 4.51  & 1.71   & \phantom{0}9.31  & 2.05  & 0.74  & 6.14 \\ 

Masked Transf.~\citep{zhou:2018} 
& \cmark & \xmark & \cmark & 5.76  & 2.71   & 11.16 & 2.91  & 1.44  & 6.91 \\

Bi-SST~\citep{wang:2018}         
& \cmark & \xmark & \cmark & $-$   & $-$    & 10.89 & 2.27  & 1.13  & 6.10 \\

SDVC~\citep{mun:2019}            
& \cmark & \xmark & \cmark & $-$   & $-$    & $-$   & $-$   & $-$   & 6.92 \\ 

\midrule

MMWS~\citep{rahman:2019}         
& \xmark & \xmark & \xmark & 3.04  & 1.46   & \phantom{0}7.23  & 1.85  & 0.90  & 4.93 \\

BMT~\citep{iashin:2020b}         
& \xmark & \xmark & \xmark & 4.63  & 1.99   & 10.90 & 3.84  & 1.88  & 8.44 \\ 

iPerceive~\citep{chadha:2021}    
& \xmark & \xmark & \xmark & \textbf{6.13}  & \textbf{2.98}  & \textbf{12.27} & 2.93  & 1.29  & 7.87 \\

MDVC~\citep{iashin:2020}         
& \xmark & \xmark & \xmark & \textbf{5.83}  & \textbf{2.86}   & \textbf{11.72} & 2.60  & 1.07  & 7.31 \\

TSP~\citep{alwassel:2020}        
& \xmark & \xmark & \xmark & $-$   & $-$    & $-$   & 4.16  & 2.02  & \textbf{8.75} \\ 

\minor{PDVC~\citep{wang:2021e2e}} 
& \xmark & \xmark & \xmark & $-$      & 3.12 & 11.26      &  $-$        &  1.96       & 8.08     \\ 

\minor{GVL~\citep{wang2023learning}} 
& \cmark & \xmark & \xmark & $-$   & $-$    & $-$   & $-$   & 2.18 & 8.50     \\

\midrule \midrule 
Ours$_{\textit{RGB+Sm}}$         
& \cmark & \xmark & \xmark & 5.40  & 2.55   & 11.06 & \textbf{4.37}  & \textbf{2.42} & 8.52 \\

Ours$_{\textit{V+Sm}}$           
& \cmark & \xmark & \xmark & 5.54  & 2.64   & 11.23 & \textbf{4.57}  & \textbf{2.55} & 8.65 \\

\bottomrule
\end{tabular}
}
\label{tab:stateoftheart}
\end{table}

%% file: 5.4-ablations.tex
\subsection{Ablation Studies}

\minor{In this section, we conduct ablation studies to assess two key aspects: (i)~the performance of the semantic descriptor under different configurations, involving changes to the vocabulary and context window, and (ii)~the impact of our proposed descriptor on the MDVC method, i.e., combining it with multiple~modalities.}

\subsubsection{Semantic Descriptor Evaluation}

We evaluated the semantic descriptor for its performance on the BMT model replacing the audio signal and considering ground truth and captioning of event proposals, Figure~\ref{fig:semanticembeddingablation} (a) and~(b), and for performance of event proposals~(c).
We changed the vocabulary size $|C|$ (100, 500, 1000, 1500, and 2000) and the context window S (10s, 30s, and 60s).

\input{figures/new_ablation/semantic_descriptor_ablation}


\minor{Upon evaluating the results presented in Figure~\ref{fig:semanticembeddingablation}, it is possible to note that small vocabularies (i.e., those with 100 and 500 visual words) produce inferior results than larger vocabulary sizes for all metrics (B@3, B@4, and M).
The best results were achieved using vocabularies with $1{,}000$ or $1{,}500$ visual words.
Surprisingly, the vocabulary comprising $2{,}000$ visual words suffers a significant drop in performance for all context window sizes.
We can also observe that the context window size significantly impacts the performance.
Long windows (i.e., $60$s) usually produce worse results than smaller windows with $30$s or $10$s.
The optimal results were achieved using~$10$s.}

\minor{The same conclusions are reached by analyzing the results of event generation performance, as shown in Figure~\ref{fig:semanticembeddingablation}~(c).
The best results are observed when using either a $30$s window with $1{,}000$ visual words or a $10$s window with $1{,}500$ visual words.
Thus, in the experiments conducted in this paper, the configuration with $1{,}500$ visual words and $10$s of context window was selected due to its superior performance across both~tasks.}

\subsubsection{Impact of co-occurrence similarity estimation on the MDVC method}
\label{subsec:mdvccombinedvisglove}

Motivated by the event proposal performance, we adopt the same features and validation sets from \gls*{bmt} in both our method and \gls*{mdvc} baseline.
This enables a fair comparison with \gls*{mdvc} and iPerceive~\citep{chadha:2021} \gls*{sota} methods, as there are a few differences between the filtered validation sets used to evaluate BMT and MDVC.
There are also differences in the number of frames used to extract visual features with the i3D method ($24$ frames~\citep{iashin:2020} $\times$ $64$~frames in our experiments).

\input{tables/mdvc_results}

Table~\ref{tab:mdvcperformance} shows the results of \gls*{mdvc} with the same \minor{i3D} features as \gls*{bmt} and with our temporal proposals using \textit{V~(\#1)}, \textit{V+A~(\#2)}, \textit{V+S~(\#3)} and \textit{V+A+S~(\#4)}, where \textit{V} = i3D output for \textit{RGB} and \gls*{of} streams, \textit{A}~=~audio, \textit{S}~=~speech\minor{, and \textit{Sm}~=~co-occurrence similarity}. \minor{We evaluate the same configurations including the \textit{Sm} descriptor (\textit{\#6}, \textit{\#7}, \textit{\#8}, and \textit{\#9}), and an additional scenario in which only RGB stream is employed \textit{(RGB+Sm)~(\#5)}.}
\minor{In the most challenging scenario, learned proposals, \textit{V+A+Sm} and \textit{V+Sm} outperformed any configuration without \textit{Sm} considering M and B@3-4.
Notably, our performance with $\textit{RGB+Sm}$ in learned proposals is competitive with the multi-modal~approach.}


\minor{It is evident that audio and speech exerted a more substantial influence on the ground truth proposals than on learned proposals results.
In the case of ground truth proposals, V+Sm performs better than V (M and B@3-4).
However, in all other scenarios, a slightly inferior performance is observed, possibly due to the difficulty of multi-modal training concerning the different times for optimization of each modality during the training~process.}

%% file: figures/new_ablation/semantic_descriptor_ablation.tex
\begin{figure}[!htb]
    \centering
    \resizebox{1.0\linewidth}{!}{
	\subfloat[][Captioning performance on ground truth proposals]{
        \includegraphics[height=22ex]{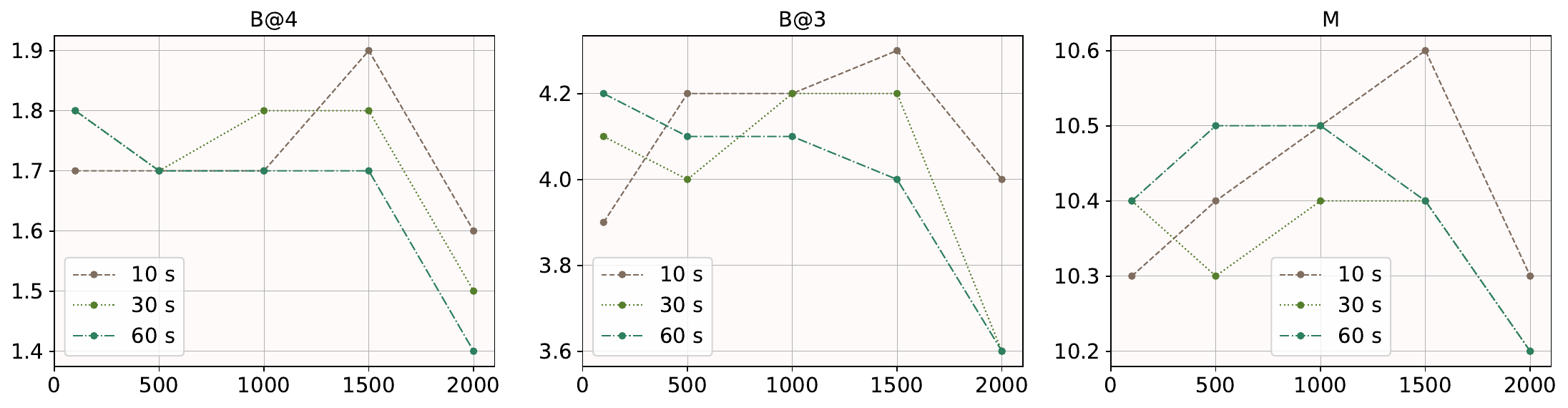} 
        \hspace{1mm}
        }
    }
    \resizebox{1.0\linewidth}{!}{
	\subfloat[][Captioning performance on event proposals]{
        \includegraphics[height=22ex]{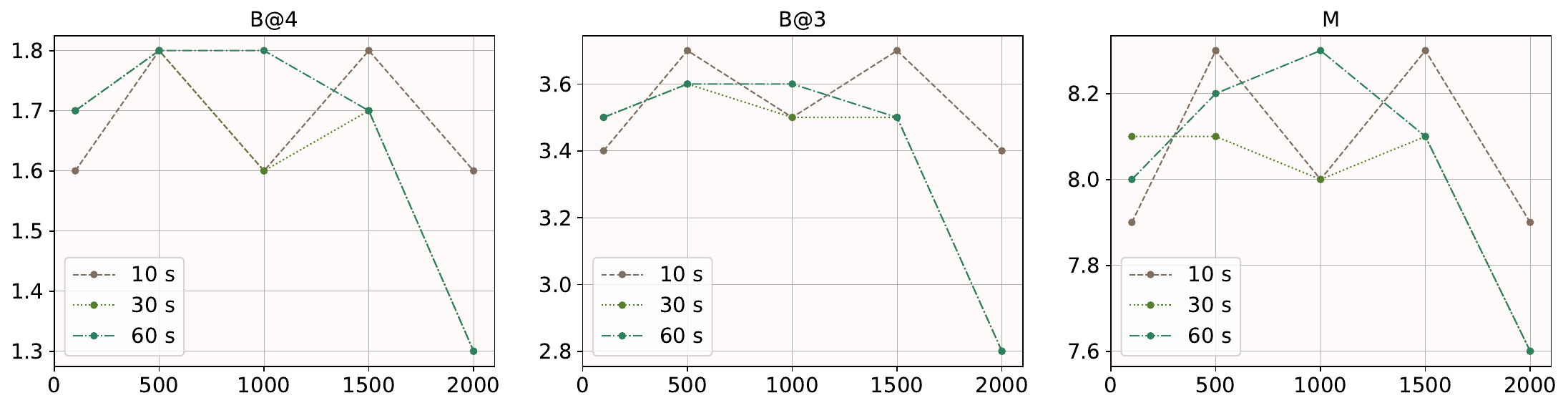} 
        \hspace{1mm}
        }
    }
    \resizebox{1.0\linewidth}{!}{
	\subfloat[][Performance of event proposals]{
        \includegraphics[height=22ex]{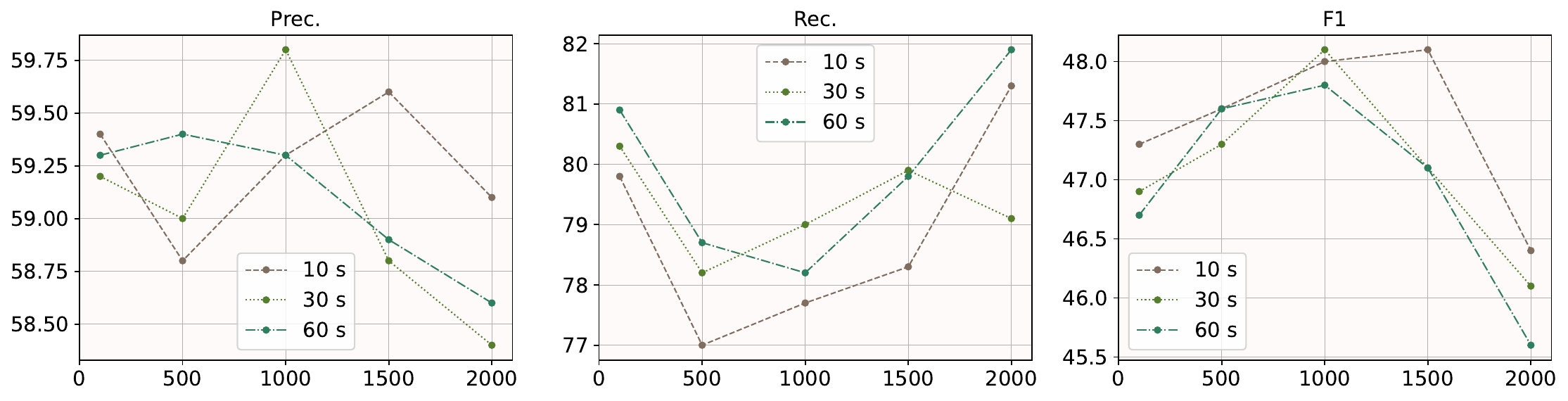} 
        \hspace{1mm}
        }
    }
    \caption{Captioning performance levels for ground truth and event proposals, (a) and (b), and for proposal generation (c) considering the \gls{bmt} model replacing the audio signal with several semantic descriptors generated with different vocabularies (100, 500, 1000, 1500, 2000) and context windows (10s, 30s, 60s).
    }
    \label{fig:semanticembeddingablation}
\end{figure}

%% file: tables/mdvc_results.tex
\begin{table*}[!htb]
\centering
\caption{Results on the ActivityNet Captions dataset~\citep{krishna:2017} adopting the MDVC method and the same validation sets used in iPerceive~\citep{chadha:2021}. \textit{V}~=~i3D output for \textit{RGB} and Optical Flow~(\textit{OF}) streams; \textit{A}~=~audio; \textit{S}~=~speech; \textit{Sm}~=~co-occurrence similarity; \textit{B}~=~BLEU@N; \textit{M}~=~METEOR; \textit{R}~=~Rouge$_{l}$; and \textit{C}~=~CIDEr-D. The top~$2$ results for each metric are highlighted in~bold.}
\vspace{-2mm}
\footnotesize
\begin{tabular}{cccccccccccccccc}
\toprule
\multirow{2}{*}{\#} & 
\multicolumn{2}{c}{$V$} & 
\multirow{2}{*}{$A$} & 
\multirow{2}{*}{$S$} & 
\multirow{2}{*}{$Sm$} & 
\multicolumn{5}{c}{GT Proposals} 
& \multicolumn{5}{c}{Learned Proposals} \\
&  {\footnotesize RGB} & {\footnotesize OF} & & &      
& B@3   & B@4  & M     & R     & C     
& B@3   & B@4  & M     & R     & C     \\ 
\midrule

1 & \cmark & \cmark  &        &        &        &             
5.40       & 2.67       & 11.18       & 22.90       & 44.49        &  
4.40       & 2.46       & 8.58        & 13.36       & \tbf{13.03}     \\ 

2 & \cmark & \cmark & \cmark &        &        & 
\tbf{5.67} & \tbf{2.75} & \tbf{11.37} & \tbf{23.69} & \tbf{46.19}  &  
4.49       & 2.50       & 8.62        & 13.49       & \tbf{13.48}     \\ 

3 & \cmark & \cmark &        & \cmark &        &             
\tbf{5.78} & \tbf{2.95} & 10.87       & 22.87       & 41.40        &  
4.21       & 2.33       & 8.43        & 13.34       & 11.79        \\ 

4 & \cmark & \cmark & \cmark & \cmark &        & 
5.61       & 2.69       & \tbf{11.49} & \tbf{23.82} & \tbf{46.29}  &  
4.41       & 2.31       & 8.50        & 13.47       & 13.09        \\ 

\midrule \midrule

5 & \cmark &        &        &        & \cmark & 
5.40       & 2.55       & 11.06       & 23.01       & 42.53        &  
4.37       & 2.42       & 8.52        & 13.40       & 12.14        \\ 

6 & \cmark & \cmark &        &        & \cmark       & 
5.54       & 2.64       & 11.23       & 23.34       & 45.76        &  
\tbf{4.57} & \tbf{2.55} & \tbf{8.65}  & \tbf{13.62} & 12.82        \\ 

7 & \cmark & \cmark & \cmark &        & \cmark & 
5.61       & 2.71       & 11.25       & 23.61       & 45.09        &  
\tbf{4.60} & \tbf{2.58} & \tbf{8.69}  & \tbf{13.87} & 12.99        \\ 

8 & \cmark & \cmark &        & \cmark &  \cmark & 
5.18       & 2.52       & 10.95       & 22.77       & 43.91        &  
3.89       & 2.07       & 8.13        & 12.78       & 11.00        \\ 

9 & \cmark & \cmark & \cmark & \cmark & \cmark & 
5.53       & 2.62       & 11.17       & 23.43       & 45.93        &  
4.10       & 2.20       & 8.30        & 13.07       & 11.64        \\ 

\bottomrule
\end{tabular}
\label{tab:mdvcperformance}
\end{table*}

%% file: 7-conclusion.tex
\section{Conclusions and Future Work}

In this work, we presented a method to enrich visual features for dense video captioning that learns visual similarities between clips from different videos and extracts information on their co-occurrence probabilities. Our conclusions are:
(i)~co-occurrence similarities combined with deep features can provide more meaningful semantic information for dense video captioning than only deep features from a single modality;
(ii)~our semantic features processed with an encoder-decoder scheme based on transformers outperformed single modality methods while achieving competitive results with multi-modal state-of-the-art methods; 
and (iii)~we reached impressive results adopting only the RGB stream when compared to results using RGB, optical flow, and audio~information.

As directions for future work, deep clustering methods could replace the mini-batch $k$-means.
As our method is unsupervised, multiple large-scale visual datasets could be combined without the need for linguistic descriptions or human annotations.
These datasets could be used to learn more accurate/detailed codebooks using co-occurrences or BERT-based~models. 
\minor{Additionally, conducting experiments on diverse domains, such as cooking activities~\citep{zhou:2018} or sport events~\citep{yu:2018}, could further enrich the scope of the~investigation.}

%% file: 9-acknowledgments.tex
\section*{{Acknowledgments}}

This work was supported by the Federal Institute of Paran\'{a},
Federal University of Paran\'{a}, and by grants from the National
Council for Scientific and Technological Development (CNPq)
(grant numbers 308879/2020-1, 304836/2022-2, and 315409/2023-1). 
The Titan Xp GPU used for this research was donated by~NVIDIA.